\documentclass[10pt,twocolumn,letterpaper]{article}

\usepackage[pagenumbers]{cvpr} 
\usepackage{comment}
\usepackage{xspace}
\usepackage{xcolor} 
\usepackage{makecell}
\usepackage{newtxtext,newtxmath}

\newcommand{\sysname}{\textsc{Envision}\xspace}

\newcommand{\image}{I}
\newcommand{\envimage}{\image_{env}}
\newcommand{\goalimage}{\image_{goal}}
\newcommand{\roiimage}{\image_{roi}}

\newcommand{\textprompt}{T}
\newcommand{\Textplanning}{\textprompt_{p}}
\newcommand{\Textobject}{\textprompt_{o}}
\newcommand{\videomodel}{\mathcal{V}}
\newcommand{\imagemodel}{\mathcal{G}}

\definecolor{cvprblue}{rgb}{0.21,0.49,0.74}
\usepackage[pagebackref,breaklinks,colorlinks,allcolors=cvprblue]{hyperref}

\def\paperID{} 
\def\confName{CVPR}
\def\confYear{2026}

\title{\textsc{Envision}: \underline{E}mbodied Visual Pla\underline{n}ning via Goal-Imagery \underline{Vi}deo Diffu\underline{sion} } 

\author{Yuming Gu$^{1,2}$ \hspace{4pt} Yizhi Wang$^{2}$\hspace{4pt} Yining Hong$^{3}$ \hspace{4pt} Yipeng Gao$^{1,2}$ \hspace{4pt} Hao Jiang$^{1}$ \hspace{4pt} Angtian Wang$^{2}$ \hspace{4pt} Bo Liu$^{2}$\\
 Nathaniel S. Dennler$^{4}$\hspace{6pt} Zhengfei Kuang$^{3}$ \hspace{6pt} Hao Li$^{5}$ \hspace{6pt} Gordon Wetzstein$^{3}$ \hspace{6pt} Chongyang Ma$^{2}$  \\
\small{$^1$University of Southern California\hspace{5pt} $^2$ByteDance \hspace{5pt} $^3$Stanford University \hspace{5pt} $^4$Massachusetts Institute of Technology \hspace{5pt} $^5$MBZUAI} \\
\small{\url{https://envision-paper.github.io/}} \\
}

\begin{document}
\definecolor{goalblue}{HTML}{4A90E2}
\definecolor{goalred}{HTML}{E74C3C}

\twocolumn[{
    \renewcommand\twocolumn[1][]{#1}%
    \maketitle
    \begin{center}
        \includegraphics[width=\textwidth]{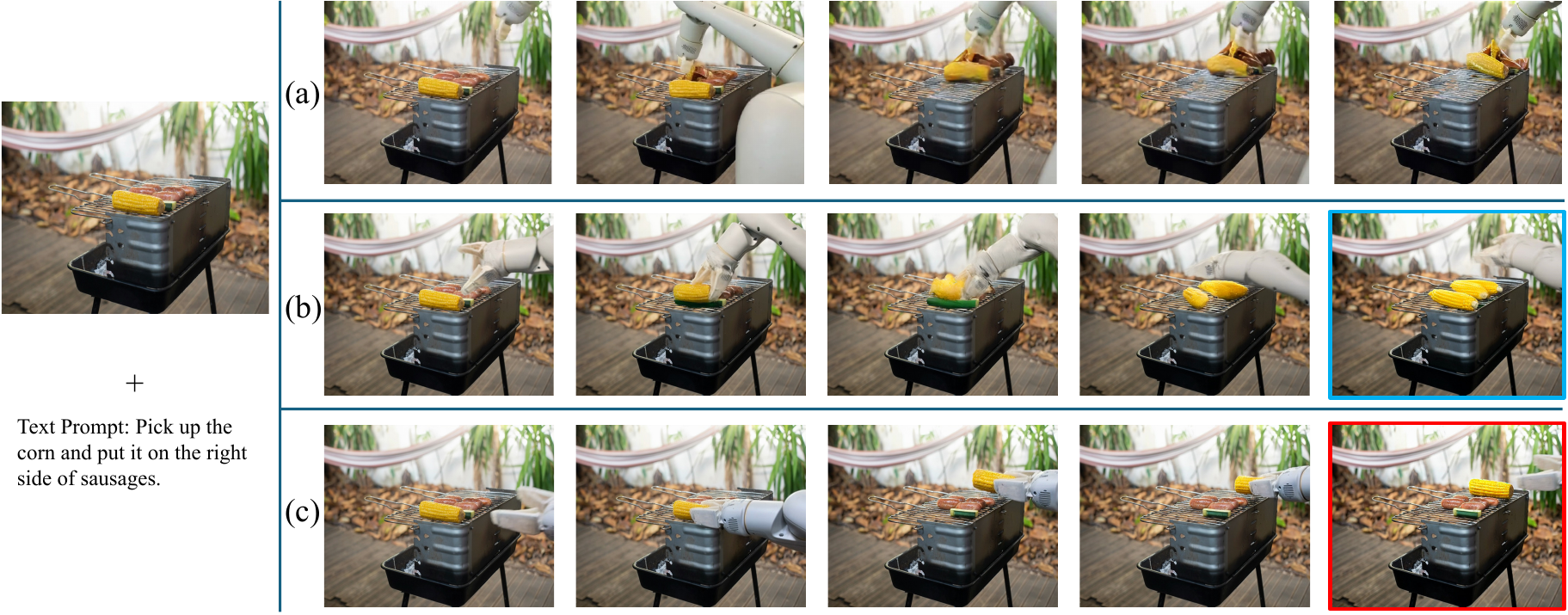}
        \captionof{figure}{Given a scene image and a task prompt, ``Pick up the corn and put it on the right side of sausages,'' (a) Tesseract~\cite{zhen2025tesseract} performs video generation from a single input image, but struggles to maintain spatial consistency and goal alignment without additional constraints. (b) A strong baseline that combines Nano Banana~\cite{google2025nanobanana} and Wan 2.1 First-Last-Frame-to-Video (FL2V) model~\cite{wan2025wan}: Nano Banana first predicts the goal (last) frame (outlined in \textcolor{goalblue}{blue}), and Wan 2.1's FL2V then synthesizes the full sequence. However, Nano Banana often causes unintended scene edits and ambiguous object identities. (c) Our \sysname accurately predicts the goal frame (outlined in \textcolor{goalred}{red}) and interpolates between the initial and goal frames, yielding a goal-aligned, spatially consistent, and physically plausible planning video.}
        \label{fig:teaser}
    \end{center}
}]

\begin{abstract}
Embodied visual planning aims to enable manipulation tasks by imagining how a scene evolves toward a desired goal and using the imagined trajectories to guide actions. Video diffusion models, through their image-to-video generation capability, provide a promising foundation for such visual imagination. However, existing approaches are largely forward predictive, generating trajectories conditioned on the initial observation without explicit goal modeling, thus often leading to spatial drift and goal misalignment. To address these challenges, we propose \textsc{Envision}, a diffusion-based framework that performs visual planning for embodied agents. By explicitly constraining the generation with a goal image, our method enforces physical plausibility and goal consistency throughout the generated trajectory. Specifically, \textsc{Envision} operates in two stages. First, a Goal Imagery Model identifies task-relevant regions, performs region-aware cross attention between the scene and the instruction, and synthesizes a coherent goal image that captures the desired outcome. Then, an Env-Goal Video Model, built upon a first-and-last-frame-conditioned video diffusion model (FL2V), interpolates between the initial observation and the goal image, producing smooth and physically plausible video trajectories that connect the start and goal states. Experiments on object manipulation and image editing benchmarks demonstrate that \textsc{Envision} achieves superior goal alignment, spatial consistency, and object preservation compared to baselines. The resulting visual plans can directly support downstream robotic planning and control, providing reliable guidance for embodied agents.
\end{abstract}



\section{Introduction}

Recent advances in generative video models~\cite{yang2024cogvideox, wan2025wan, agarwal2025cosmos,assran2025v,dai2025fantasyworld,HaCohen2024LTXVideo} have dramatically expanded the frontier of visual planning for robots, enabling robots not only to perceive but also to imagine temporal unfoldings of scene dynamics in response to potential interactions. These models, often conceptualized as learned world models \cite{jang2025dreamgen, zhen2025tesseract, barcellona2025dreammanipulatecompositionalworld, zhu2025unifiedworldmodelscoupling, guo2025flowdreamerrgbdworldmodel}, employ diffusion-based generative techniques to simulate future visual observations conditioned on current observation and action instructions, thereby providing robots with a form of visual foresight to guide their actions.

However, despite their expressive power, current video diffusion models are predominantly forward-simulative: given an initial observation and a text instruction, world models \cite{zhen2025tesseract, agarwal2025cosmos} generate videos that depict how the scene might evolve under the text instruction, without explicit guidance from a desired goal. Consequently, in the absence of end-state constraints, the denoising process lacks the necessary information to anchor the generation toward the desired goal, leading to noticeable drift, deformation, and other physically inconsistent artifacts in the generated frames. Eventually, these inconsistencies lead to \emph{physically implausible} and \emph{goal-misaligned} video predictions, as illustrated in Figures \ref{fig:teaser} and \ref{fig:comparison_with_sota_video_mdoels}. Such deviations in the generated trajectories substantially degrade downstream robotic execution performance, frequently leading to task failures, as exemplified in Figure \ref{fig: rollbot simulation}.

\begin{figure*}
  \centering
  \includegraphics[width=0.98\textwidth]{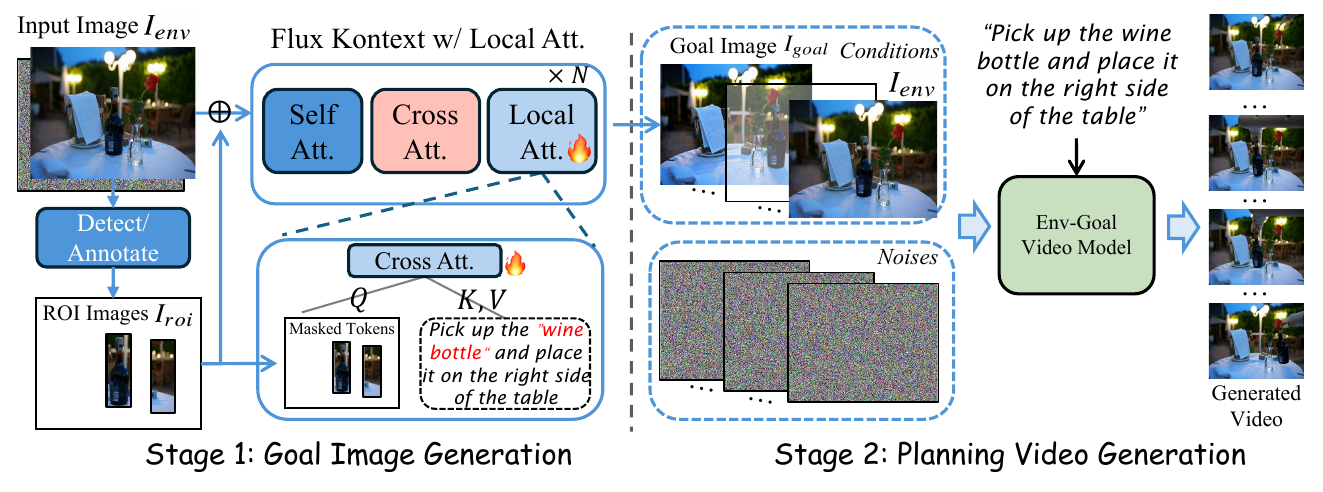}
  \caption{Overview of our \sysname framework.
  Given a single environment image and an instruction prompt, our pipeline generates a physically plausible and goal-aligned video depicting the instructed manipulation in a two-stage manner. Each stage corresponds to a trainable component: (left) a Goal Imagery Model that predicts the target goal frame, and (right) an Env–Goal Video Model that synthesizes the full sequence conditioned on both the environment and goal images. 
  }
  \label{fig:pipeline}
\end{figure*}

Humans, in contrast, plan by first envisioning the goal and working backward from it. When presented a task such as ``Pick up the wine bottle and place it on the right side of the table'', as shown in Figure \ref{fig:pipeline}, humans typically first envision the desired final state, with the bottle properly positioned on the table's right side, and then mentally simulate the transformations required to achieve the goal. This form of goal-directed cognition constrains imagination by the desired outcome, enabling humans to reason backward from what should happen rather than forward from what might have happened. Therefore, it is crucial to establish a precise representation of the end state to guide goal-directed backward planning.

The challenge now arises: how can a robot form such a precise, goal-aligned visual representation? One natural solution is to leverage image editing techniques to modify the first frame toward the desired outcome. However, existing image editing models predominantly rely on global conditioning, where text prompts have a strong influence to the entire image, which often produces unintended edits and ambiguity, as the model may misinterpret which object the instruction refers to, \eg, referencing the ``corn'' could lead to ambiguous and undesirable result as shown in Figure \ref{fig:teaser}(b). While such global edits work for stylistic or aesthetic changes, they are ill-suited for embodied tasks that require precise, fine-grained reasoning. To enable goal-directed planning and reliable robot execution, goal generation must therefore be local, compositional, and object-centric, targeting only the entities relevant to the task while preserving the rest of the scene.

To this end, we propose \sysname, a video diffusion framework designed to generate precise and goal-aligned visual representations that facilitate effective planning. The framework consists of two stages. In the first stage, we introduce the \emph{Goal Imagery Model} that grounds textual instructions to identify task-relevant regions in the scene. It then leverages global scene context and object-specific latent features to generate a coherent goal image, implementing the desired edits in a localized and compositional manner. These edits modify only task-relevant regions while preserving the overall spatial and relational structure of the scene, thereby ensuring consistency across objects and unaffected areas. By jointly reasoning over both individual objects and the broader scene context, the framework is able to generate detailed goal images that capture the intended arrangement and interactions among all relevant elements.

In the second stage, we propose the \emph{Env-Goal Video Model}, which differs from conventional image-to-video (I2V) methods~\cite{wan2025wan, blattmann2023stable, yang2024cogvideox} that generate subsequent frames from a single input image. Instead, our model generates intermediate frames that smoothly interpolate between the initial state and the target configuration. By propagating visual transformations implied by the goal image backward in time, the model produces a continuous and visually coherent video trajectory. This interpolation not only completes the intermediate states in a physically plausible and visually consistent manner, but also captures object interactions and scene dynamics as evidenced in Figure \ref{fig:comparison_with_sota_video_mdoels}. The resulting video serves as a rich and goal-conditioned representation that can be directly leveraged for robotic planning as shown in Figure \ref{fig: rollbot simulation}, providing a detailed roadmap from the current scene to the target configuration. Furthermore, to ensure both realistic motion synthesis and generalization to robotic settings, the model is trained with a cross-modal coherence strategy, leveraging a mixture of human-hand and robotic manipulation videos.
While prior video diffusion models are primarily trained on I2V settings, our Env-Goal Video model enables the learning of precise planning dynamics and physically consistent transformations, from a diverse range of cross-domain datasets.

Our contributions can be summarized as follows:
\begin{itemize}
\item We introduce \sysname, a two-stage framework for embodied visual planning that generates precise, goal-aligned video trajectories.
\item We propose the Goal Imagery Model, integrating a goal-a ware attention mechanism which identifies task-relevant regions and constructs coherent goal images in a localized and compositional manner.
\item We develop the Env-Goal Video Model, which interpolates intermediate frames to produce smooth and physically plausible video trajectories.
\item Through experiments on object manipulation and image editing benchmarks, we demonstrate that \sysname generates task-aligned videos that preserve spatial and relational consistency, while supporting downstream robotic planning.
\end{itemize}

\section{Related Work}
\label{sec:related_work}

\paragraph{Image generation and editing models.}
Existing visual manipulation and embodied planning approaches across diverse research paradigms~\cite{ebert2018visual, shah2021ving, monaci2025does, goodwin2022semantically} have demonstrated that using a goal image is effective in improving system performance.
However, in practical embodied task scenarios, real-world goal images are typically unavailable.
With the recent significant advances in GANs~\cite{goodfellow2014generative} and diffusion models~\cite{ho2020denoising,sd2022}, the generation of goal images has become feasible.
While GAN-based methods such as StyleCLIP~\cite{patashnik2021styleclip} and StyleFlow~\cite{abdal2021styleflow} enable text-driven image editing by leveraging CLIP~\cite{radford2021learning} features, they often suffer from limited latent expressivity and struggle to model the complex relationships between text prompts and images.

Diffusion models~\cite{ho2020denoising} have emerged as one of the most powerful alternatives for image generation.
Early studies introduce diffusion-based image editing via cross-attention manipulation or inversion~\cite{hertz2022prompt, mokady2023null, wallace2023edict}.
InstructPix2Pix~\cite{shagidanov2024grounded} further shifts the paradigm toward user-friendly, single-instruction editing.
Editing quality continues to advance with flow-matching models such as Flux Kontext~\cite{batifol2025flux} and closed-source systems such as Nano Banana~\cite{google2025nanobanana}.
Furthermore, OmniGen2~\cite{wu2025omnigen2} and BAGEL~\cite{deng2025emerging} integrate large vision–language models with generative decoders to enable context-aware editing.
Compared to GAN-based methods, diffusion-based image editing models are trained on larger-scale datasets and incorporate reasoning abilities, enabling richer and more controllable manipulations in both appearance and semantics.

\paragraph{Video generation and world models.}
Recent advances in diffusion models have also substantially improved video generation quality.
Early approaches~\cite{du2023learning,yang2023learning} typically employ U-Net-based latent video generation models~\cite{ho2022imagen} for the generation of observation sequences. With progress in video foundation models, recent work~\cite{wang2025language, luo2025solving, jang2025dreamgen, zhen2025tesseract, dai2025fantasyworld, assran2025v} has achieved more powerful and visually consistent results by applying DiT-based video models~\cite{yang2024cogvideox, HaCohen2024LTXVideo, genmo2024mochi, agarwal2025cosmos, wan2025wan}. While trained on billions of videos, video models can already generate photorealistic appearances that are nearly indistinguishable from real footage, but they still fail to fully comprehend the underlying physical laws governing the real world. As a result, a number of methods~\cite{chefer2025videojam, zhao2025synthetic, yuan2025newtongen, shaulov2025flowmo} are working on improving the physical consistency of foundation models. Though those methods bring measurable improvements to general-purpose text-to-video or image-to-video generation, embodied visual planning requires a higher standard, as it requires generating physically consistent and instruction-aligned trajectories.

On the other hand, a series of methods~\cite{jang2025dreamgen, luo2024grounding, zhen2025tesseract, assran2025v, guo2025ctrl} working on embodied visual planning aim to follow text instructions and generate physically plausible results. Tesseract~\cite{zhen2025tesseract} introduces depth and normal information for a 4D embodied world model. V-JEPA 2~\cite{assran2025v} enhances physical understanding by learning object-centric latent dynamics that predict how scenes evolve over time, enabling the model to internalize causal and consistent motions and interactions directly from video data. This \& That~\cite{luo2024grounding} introduces additional points to reduce ambiguities, though it may lead to shape inconsistencies during planning. While all these methods leverage different priors to make robot planning more precise, none of them can achieve goal alignment while maintaining physical consistency, which is crucial for the success of a visual planning system.

\section{Method}
\label{sec:method}

Given an RGB environment image $\envimage$ and a textual instruction $T_{p}$, our objective is to generate a sequence of frames $\{ \image_{p_t} \}_{t=1}^{T}$ that visually depict a coherent execution of the instructed plan. The generated sequence should preserve the visual fidelity and spatial structure of the original environment in $\envimage$, while faithfully illustrating the agent's embodied reasoning and actions as specified by the instruction.

As illustrated in Figure~\ref{fig:pipeline}, our \sysname follows a two-stage generation paradigm.
We first present the preliminaries in Section~\ref{sec:preliminaries}.
Next, we introduce the Goal Imagery Model (Section~\ref{sec:goal_imagery_model}), which generates a realistic and spatially consistent goal image $\goalimage$ from the input while mitigating unrealistic scene edits and focusing on the most relevant regions of interest. 
Building on this foundation, we propose the Env-Goal Video Model (Section~\ref{sec:fl2v_video_model}), which generates the planning video, \ie, an image sequence $\{ \image_{p_t} \}_{t=1}^{T}$, explicitly conditioned on both the environment frame $\envimage$ and the predicted goal frame $\goalimage$.
Finally, Section~\ref{sec:method_training} describes a hybrid training strategy that integrates cross-domain robot and human-hand datasets to improve generalization across embodiments while maintaining both visual and behavioral consistency.


\subsection{Preliminaries}
\label{sec:preliminaries}

\paragraph{Image editing models.}
Diffusion models~\cite{song2020denoising,song2020score,ho2020denoising} demonstrate remarkable capability in learning complex data distributions $p(x)$ by progressively adding Gaussian noise to data samples through a forward process until the signal becomes indistinguishable from pure noise. During inference, a trained denoising network is applied to iteratively reconstruct the data by reversing this process.
Building upon this foundation, image editing models~\cite{batifol2025flux, shi2024seededit} are generative models designed to modify a given image according to condition inputs such as text prompts, while preserving the original image identity and structure. Our last-frame generation framework builds on the Latent Diffusion Model~\cite{sd2022} paradigm, operating in the compressed latent space of a pretrained VAE, which encodes condition images at adaptive resolutions into compact visual latents. For our last-frame generation backbone, we adopt the DiT-based Flux Kontext Model~\cite{batifol2025flux}. It employs flow matching~\cite{lipman2022flow} as the training objective, enabling efficient learning of the reverse denoising process.

\paragraph{Video diffusion models.}
Different from image editing models, latent video diffusion models~\cite{ho2022imagen,wan2025wan, yang2024cogvideox, blattmann2023stable} incorporate a Variational Autoencoder (VAE)~\cite{van2017neural} to encode video data into a compact latent representation, preserving high-fidelity visual details while improving the efficiency of modeling the rich spatio-temporal dynamics inherent in video generation tasks.
Our model is built upon the pretrained Wan backbone~\cite{wan2025wan}, which adopts a flow-matching~\cite{lipman2022flow} diffusion formulation conditioned on both the initial image and textual prompt.


\subsection{Goal Imagery Model}
\label{sec:goal_imagery_model}

\begin{figure}
  \centering
  \includegraphics[width=\columnwidth]{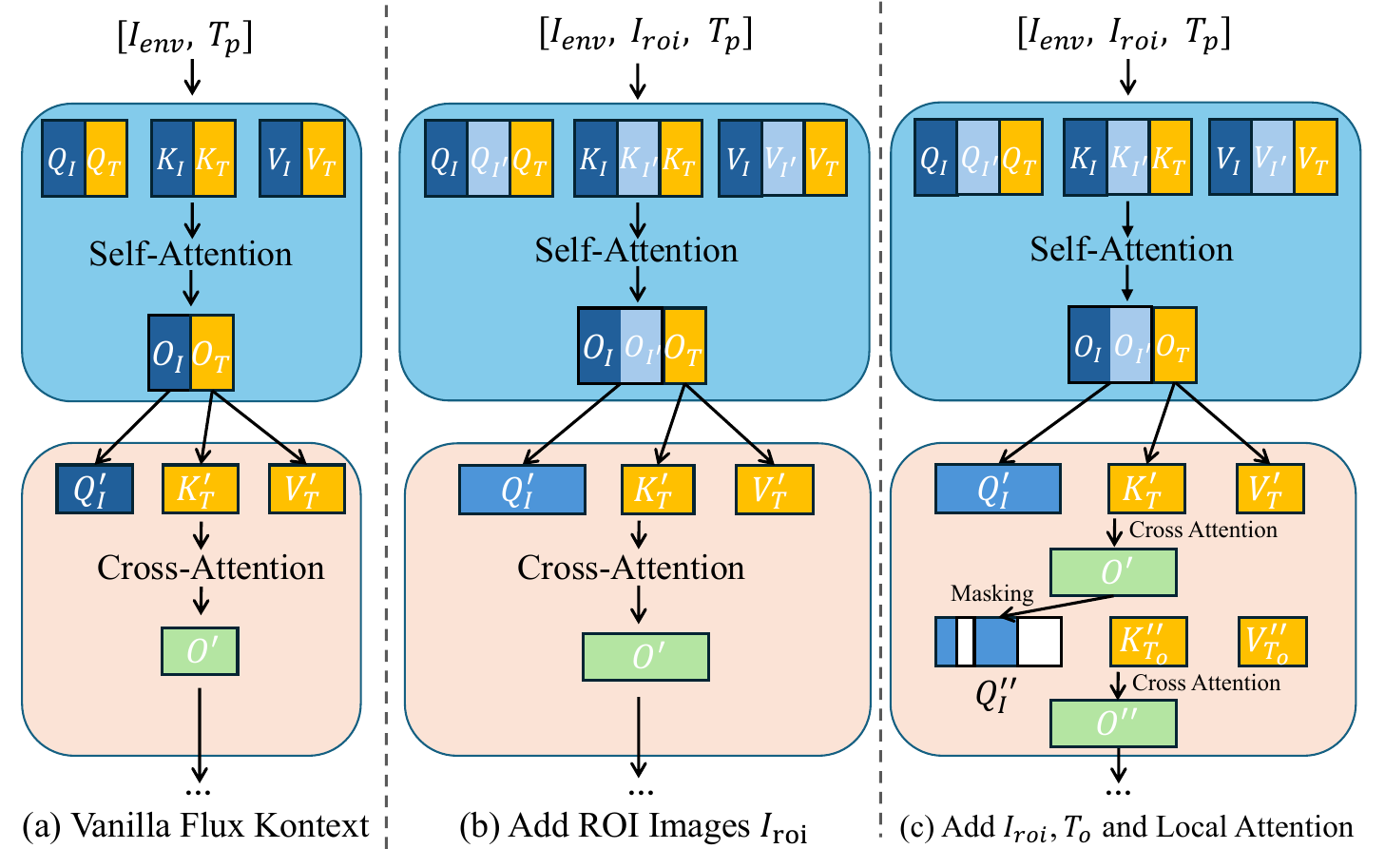}
  \caption{Our goal image generation pipeline builds on (a) Flux Kontext \cite{batifol2025flux}, which interleaves self- and cross-attention between visual and textual tokens. To better guide the model's focus toward regions of interest (ROIs), we extend the input with the corresponding ROI image $\roiimage$ in addition to the environmental image $\envimage$, as shown in (b). To further specify where textual instructions apply within $\envimage$, (c) adds an extra gated cross-attention layer between textual tokens and masked ROIs, termed Local Attention, on top of (b).}
  \label{fig:goal_image_pipeline}
\end{figure}

\begin{figure}
  \centering
  \includegraphics[width=1.0\linewidth]{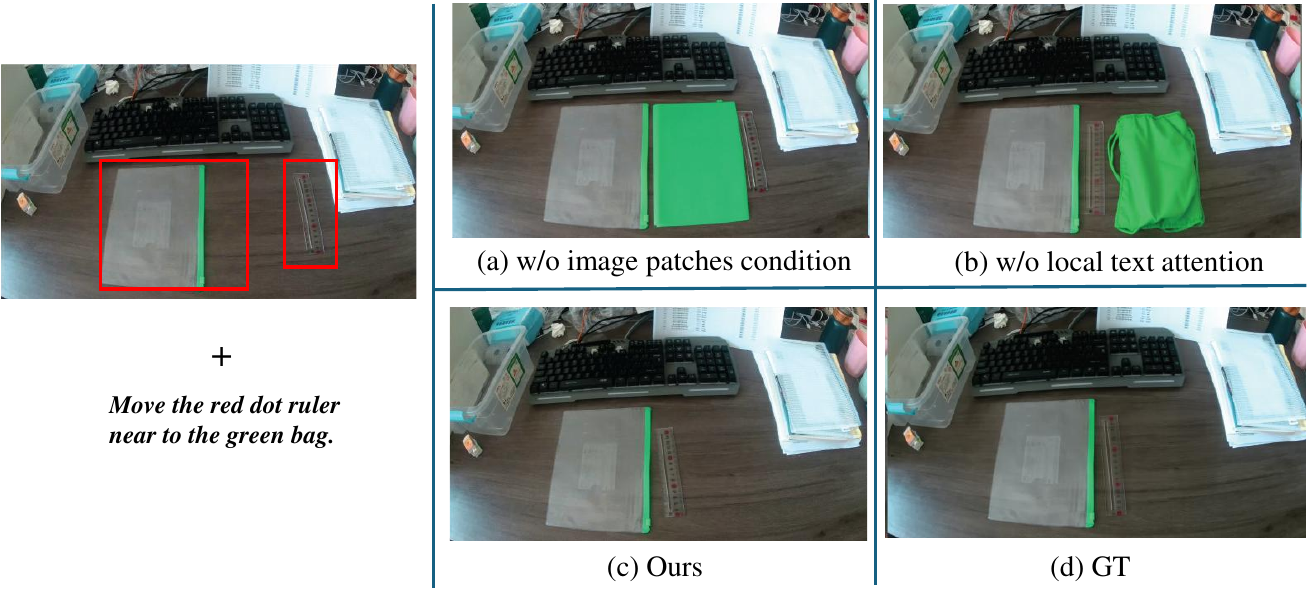}
  \caption{Ablation on our Goal Imagery Model.
  Without our local mask and local text conditions, global text conditioning tends to misinterpret the instructions. When only the local mask condition is applied, unintended artifacts and objects appear. In contrast, our local mask–text attention helps maintain structural coherence and more precisely follow the edit instructions.
  }
  \label{fig:goal_image/ablation/}
  \label{fig:goal_image_ablation}
\end{figure}

For embodies planning tasks, it remains challenging to obtain a reasonable goal image $\goalimage$, especially when it differs drastically from $\envimage$ (\eg, deformable objects).
To address this issue, we aim to gain control over the synthesized goal image without influencing either the derived appearance or the planning capability of the pretrained video model. This naturally leads to the paradigm of Flux Kontext~\cite{batifol2025flux}, where the goal is to edit an environment image $\envimage$ based on a guidance text prompt.
Here we denote our image editing model as $\imagemodel$, to be trained with paired $(\envimage, \, \goalimage)$ images.
One straightforward design of $\imagemodel$ could directly apply the text prompt to full cross-attention.
However, we argue that such a design will introduce unwanted artifacts, such as incorrect or unwanted objects, which are likely to be propagated by the video generation model $\videomodel$ (see Section~\ref{sec:fl2v_video_model}).
Herein, attention to the whole image will be reflected in other regions of the environment image during inference. This artifact is more pronounced when the text prompt does not precisely describe objects (see Figure~\ref{fig:goal_image_ablation}).

\begin{figure}
  \centering
  \includegraphics[width=1.0\linewidth]{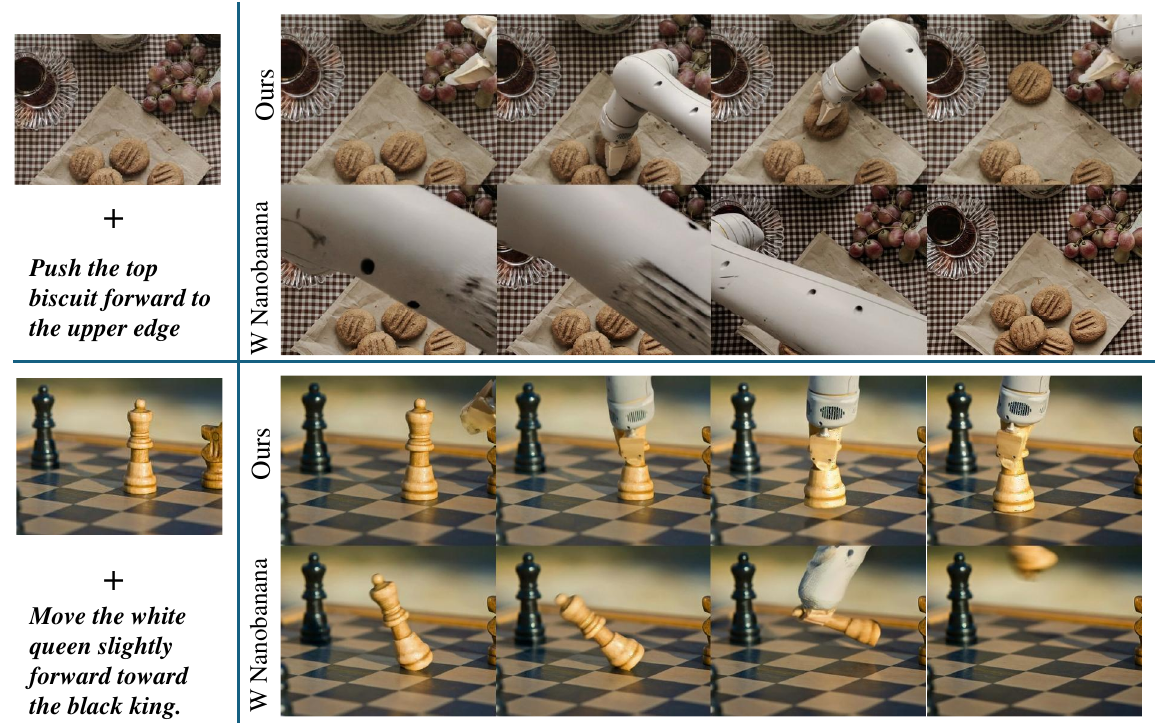}
  \caption{Comparison of video generation results using goal images produced by Nano Banana~\cite{google2025nanobanana} and \sysname. Our method generates goal images that effectively maintain physical consistency and goal alignment, resulting in more coherent and physically plausible video planning sequences.}
  \label{fig:nanobanana_video_compare}
\end{figure}

Instead, we employ a local region attention mechanism that applies text-guided attention to the region-of-interest area $\roiimage$ (as illustrated in Figure~\ref{fig:goal_image_pipeline}). This design naturally facilitates disentanglement between the background and the object of interest, enabling more precise manipulation. However, training such local region attention for goal-image generation requires paired image regions, and obtaining such data pairs is usually challenging. To address this hurdle, we leverage a state-of-the-art LLM~\cite{achiam2023gpt} to extract the object prompt, $\Textobject$, from text and then ground the object area with the object prompt, following~\cite{liu2024grounding}. This region is further enhanced through a paired low-level vision filter, VLM~\cite{achiam2023gpt}, and human-in-the-loop validation. We decompose training into two stages. We first concatenate region-of-interest (ROI) features with the environment image features to enforce localized editing (Figure~\ref{fig:goal_image_pipeline}(b)), and apply LoRA~\cite{hu2021lora} finetuning. In the second stage, in order to maximally preserve the base model’s editing behavior, a local-attention module is introduced to locally align object-specific textual information with ROI area (Figure~\ref{fig:goal_image_pipeline}(c)), allowing more precise and localized control over object appearance and behavior while keeping the rest of weights frozen.
Our Goal Imagery Model can be formulated as:
\begin{equation}
\goalimage =
\imagemodel_{\theta}\big(
\envimage,
\roiimage;
\Textobject,
\Textplanning
\big),
\end{equation}
where $\imagemodel_{\theta}$ denotes our goal-imagery editing network parameterized by $\theta$, $\roiimage$ is the localized region-of-interest mask that constrains cross-attention to the object area, and $\Textobject$ represents the object-specific instruction extracted from $\Textplanning$. Our designed local-patch attention enhancer, combined with object text attention, greatly reduces the complexity of video generation by precisely locating the regions of interest and preventing the model from creating unintended artifacts, which would otherwise impose additional burdens on the video generation process (see Figure~\ref{fig:nanobanana_video_compare}).

\subsection{Env-Goal Video Model}
\label{sec:fl2v_video_model}

To ensure generating simulated planning videos, previous methods~\cite{zhen2025tesseract,jang2025dreamgen} typically employ I2V tasks to inject the environment image and guide the denoising process with $\envimage$ and $\Textplanning$ in order to generate planning information. The I2V models are trained to ensure meticulous transfer of environment structure and appearance into the DiT blocks~\cite{peebles2023scalable} for the task of planning embodied daily tasks. While text-based goal specification can be effective for planning daily tasks, there are two key limitations: First, the goal success rate largely depends on the quality and clarity of the text prompts. When a scene contains multiple objects, the text prompt alone often fails to effectively depict the intended instructions due to prompt ambiguity. Ambiguities in the text prompt can cause compounding errors that accumulate during the forward generation process.
Second, the lack of explicit conditioning on the final state leads to varying degrees of object deformation and, in some cases, the emergence of undesirable new objects (as shown in Figure~\ref{fig:comparison_with_sota_video_mdoels}).
Such inconsistencies further compromise the accuracy of the planned outcomes. As a result, the system tends to fail when generating sequences conditioned solely on the environment image and text prompt.

\begin{figure}
  \centering
  \includegraphics[width=0.48\textwidth]{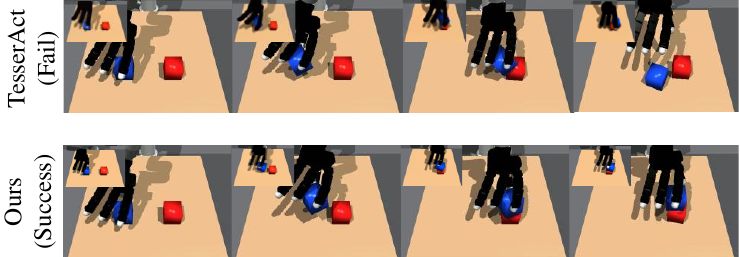}
  \caption{TesserAct~\cite{zhen2025tesseract} produces visually plausible but physically inconsistent trajectories, causing object drift and execution failure. In contrast, our method generates stable, goal-aligned video plans that lead to successful robot execution. The top-left image represents frames from the generated planning video.}
  \label{fig: rollbot simulation}
\end{figure}

\begin{figure}
  \centering
  \includegraphics[width=1.0\linewidth]{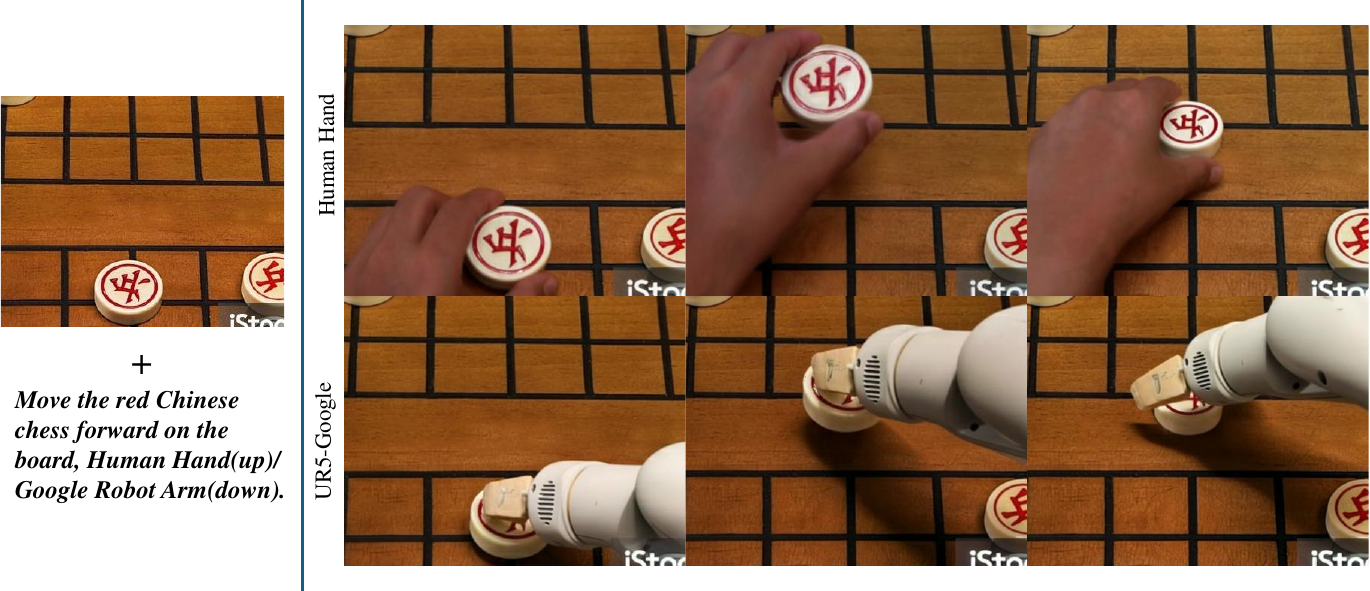}
  \caption{Our method produces consistent and goal-aligned trajectories across human hands and a UR5/Google robot arm, demonstrating strong cross-embodiment generalization.}
  \label{fig:cross_embodiement}
\end{figure}

To mitigate these issues, we develop a video diffusion model, termed the Env-Goal Video Model $\videomodel$, which incorporates information from both the start and end states to generate planning videos. This design enables the diffusion network to generate intermediate steps while preserving its full expressivity.
Specifically, $\videomodel$ requires two key inputs: the environment image $I_{\text{env}}$, representing the initial frame, and the goal image $I_{\text{goal}}$, representing the desired end state. We note that when selecting $\envimage$ and $\goalimage$, the chosen images should ideally cover the full semantic range implied by the guiding text prompt. Consequently, during LoRA finetuning~\cite{hu2021lora} on~\cite{wan2025wan}, to ensure that the text prompt best aligns with the training video semantics, we use the ground-truth last frame as $\goalimage$, since it most comprehensively captures the range described by the text prompt. As a result, we observe consistent and satisfactory planning videos that follow our demand. This makes our model $\videomodel$ capable of generating consistent planning information and potentially benefiting robot planning, as shown in Figure~\ref{fig: rollbot simulation}. The formulation can be described as follows:
\begin{align}
\resizebox{0.92\hsize}{!}{$
\mathcal{L}_{\text{FL2V}} =
\mathbb{E}_{x_0,\, t \sim \mathcal{U}(0,1),\, \epsilon \sim \mathcal{N}(0,I)}
\Big[
\big\|
v_\theta(x_t, t \,|\, \envimage, \goalimage, \Textplanning)
- (x_0 - x_t)
\big\|_2^2
\Big],
$}
\end{align}
where \(x_t = (1 - t)x_0 + t\,\epsilon\) defines the interpolated latent at continuous time \(t\),  
and \(v_\theta\) represents the learned velocity field within the DiT backbone~\cite{peebles2023scalable}.  
By jointly conditioning on $(\envimage, \goalimage, \Textplanning)$, the model explicitly preserves both environmental context and goal semantics throughout the generation trajectory, alleviating deformation or hallucination issues caused by ambiguous text-only conditioning.

\subsection{Mixed Data Training for Cross-Domain Video Planning}
\label{sec:method_training}

To this end, we have already been capable of generating planning videos following a precise text prompt by design. Nevertheless, training only on videos with a human-hand dataset limits the motion-planning sequences in the generation ability of the Env-Goal Video Model. On the other hand, the human-hand dataset further limits the model's ability in downstream applications such as robot planning, as there exists a drastic domain gap. We therefore introduce a mixed-data training strategy, facilitating the video diffusion backbone to learn both human and robot planning priors. Specifically, we integrate real human planning videos~\cite{zhao2025taste} alongside one extra robot datasets~\cite{brohan2022rt} for training. By using this mixed data, our model not only achieves human-hand-realistic planning videos, but also generalizes well to different robot arms, enabling downstream applications for robot manipulation and planning, as illustrated in Figure~\ref{fig:cross_embodiement}.

\begin{table}
\centering
\resizebox{\linewidth}{!}{
\begin{tabular}{l|l|ccccc}
\toprule
 Dataset & Method & FVD$\downarrow$ & LPIPS$\downarrow$ & PSNR$\uparrow$ & \makecell{PA$\uparrow$} & \makecell{IF$\uparrow$} \\

\midrule
 
          & Taste-Rob-I2V \cite{zhao2025taste}   & 33.04 &0.48&14.94  &65\% &32\%\\
Taste-Rob &Wanx-2.1-I2V*        &  12.95  & 0.26 &19.50 &64\% &25\% \\
          & Ours              & \textbf{8.21}  & \textbf{0.13} & \textbf{22.90} &\textbf{78\%} &\textbf{67\%} \\

\midrule
    & TesserAct \cite{zhen2025tesseract}       & 16.26  &0.29 & 16.63 &33\% & 51\% \\
RT1 &Wanx-2.1-I2V*    &  30.59 & 0.46 &14.97  &22\% &16\% \\
&Ours              & \textbf{9.95}  & \textbf{0.17}&\textbf{20.46} &\textbf{61\%} & \textbf{54\%}\\
\bottomrule
\end{tabular}
}
\caption{Quantitative comparison of generated planning videos. PA denotes physical alignment and IF denotes instruction following metrics conducted in \cite{jang2025dreamgen}, we evaluate Wanx-I2V~\cite{wan2025wan} finetuned (denote as *) in both Taste-Rob and RT1 dataset for fair comparsion. For RT1, we use the first 100 videos and Qwen2.5-VL for evaluation.}
\label{tab:quant-video-table}
\end{table}

\section{Experiments}

\subsection{Experimental Setup}

\begin{figure*}
  \centering
  \includegraphics[width=\textwidth]{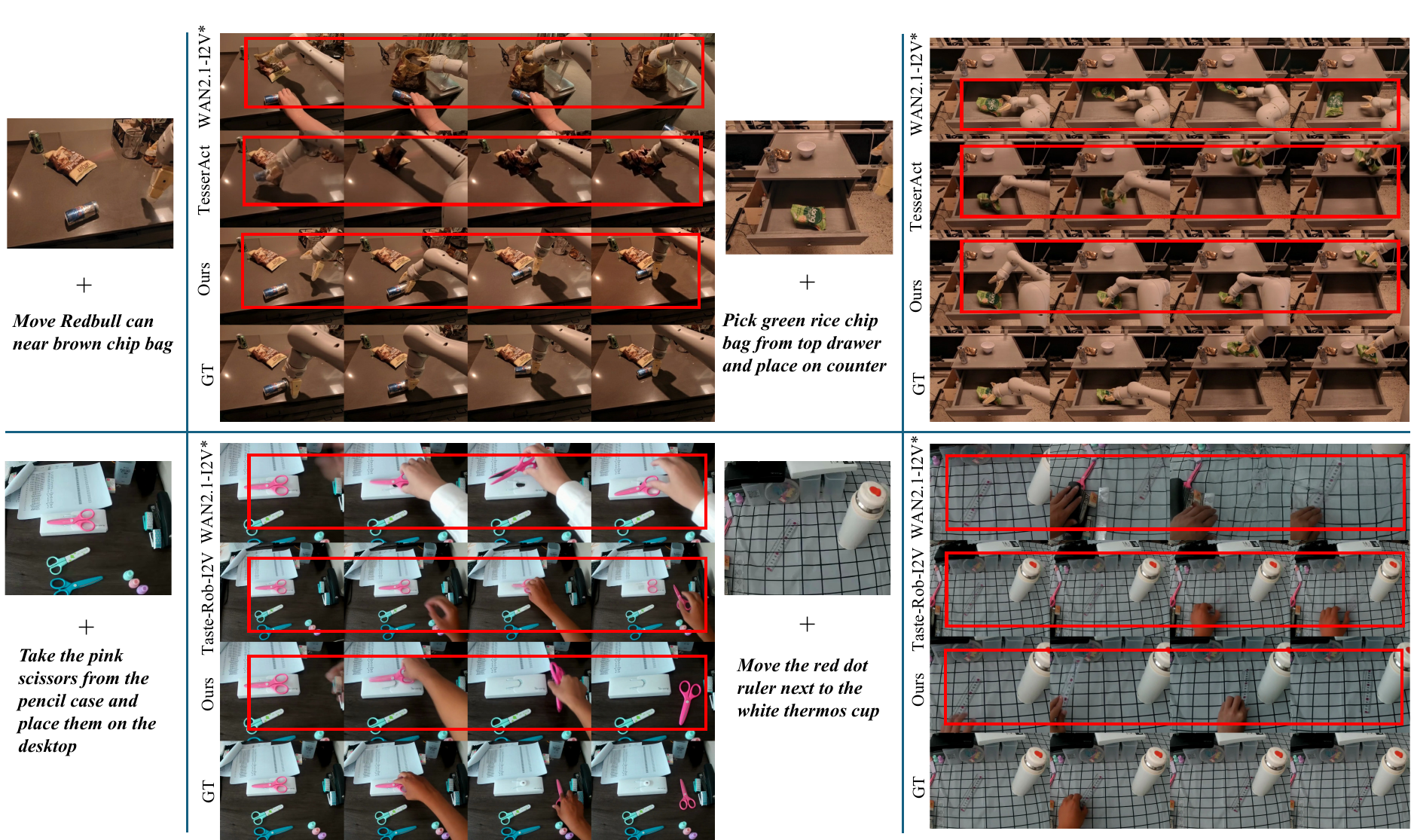}
  \caption{Qualitative comparison of video planning results on Taste-Rob~\cite{zhao2025taste} and RT-1 \cite{brohan2022rt}. Our method produces physically plausible and goal-consistent videos under diverse motion trajectories and object interactions. Compared with I2V-based baselines, \ie, Wan-2.1-I2V*, TesserAct, and Taste-Rob-I2V, our method better preserves object shape consistency and produces realistic dynamics that more closely reflect feasible robot executions.}
  \label{fig:comparison_with_sota_video_mdoels}
\end{figure*}





\vspace{-1mm}

\paragraph{{Datasets.}}
We evaluate our methods on three different datasets: Taste-Rob \cite{zhao2025taste},  RT1 \cite{brohan2022rt}, and a mixed dataset that we construct from IsaacGym \cite{makoviychuk2021isaac} and robomimic \cite{mandlekar2021matters}. For Taste-Rob, we randomly sample 200 examples spanning diverse tasks (e.g., pick and place, fold and unfold, move and push) and environments. For RT1, we select 100 different unseen real-domain examples. For the mixed dataset, we combine five tasks from~\cite{mandlekar2021matters} with six additional self-collected challenge tasks from IssacGym~\cite{makoviychuk2021isaac}. 


\begin{table}
\centering
\resizebox{\linewidth}{!}{
\begin{tabular}{l|ccccccc}
\toprule
\makecell[c]{Method} & 
\makecell[c]{Block\\sorting} & 
\makecell[c]{Block\\stack} & 
\makecell[c]{Cup\\pour} & 
\makecell[c]{Grasp\\coconut} & 
\makecell[c]{Place\\apple} & 
\makecell[c]{Soccer\\goal} \\
\midrule
Tesseract \cite{zhen2025tesseract}& 91 & 0 &0&86&28&38\\
Wan-2.1-I2V* \cite{wan2025wan}             & 0 & 0 &0&15&100&39 \\
Ours              & \textbf{100}&\textbf{34} &\textbf{87}&\textbf{100}&\textbf{100}&\textbf{50}\\
\bottomrule
\end{tabular}
}
\caption{Quantitative comparison of robot execution performance. Each method generates input frames, from which robot actions are extracted using \cite{chi2025diffusion} and executed on the robot.}
\label{tab:quant-robot_rollout}
\end{table}

\paragraph{{Metrics.}}
We evaluate our method along three dimensions: (1) perceptual and physical quality of video generation, (2) goal-image generation quality, and (3) success rate in robot simulation. For goal-image quality, we report SSIM~\cite{wang2004image}, PSNR, and LPIPS~\cite{zhang2018unreasonable} computed over the full image. For video generation quality in both the Real-Hand and Robot-Arm domains, we report FVD~\cite{unterthiner2019fvd}, LPIPS~\cite{zhang2018unreasonable}, PA~\cite{jang2025dreamgen}, and IF~\cite{jang2025dreamgen} to assess perceptual fidelity and physical plausibility. Finally, for robot experiments, we measure task performance by the success rate over 100 rollouts across a diverse set of challenge tasks.

\paragraph{{Implementation details.}}
Each training video is an action sequence consisting of 81 frames. We train our goal-image generation model follows the same training strategy as Flux Kontext~\cite{batifol2025flux}. For the robot execution policy, we train each task using 50 randomly sampled trajectories following~\cite{chi2025diffusion}. More implementation details are provided in the supplementary material.

\begin{table}
\centering
\resizebox{\linewidth}{!}{
\begin{tabular}{l|cccc}
\toprule
Method            & LPIPS$\downarrow$  & PSNR$\uparrow$ & SSIM$\uparrow$ \\
\midrule
NanoBanana \cite{google2025nanobanana}                & 0.18/0.32 & 21.05/16.43 & 0.69/0.32 \\
Kontext Flux  \cite{batifol2025flux}           & 0.34/0.37 & 15.81/15.26 & 0.38/0.48 \\
\midrule
w/o ROI images       & 0.10/0.22 & \textbf{24.79}/18.32 &0.82/0.78 \\
w/o local attention       & 0.10/0.21 & 24.08/18.08 &0.81/0.77 \\
Ours                       & \textbf{0.09/0.20} & 24.65/\textbf{18.91}  &\textbf{0.83/0.77} \\
\bottomrule
\end{tabular}
}
\caption{Quantitative-comparison of goal image generation network against SOTA methods and ablation analysis on the Taste-Rob~\cite{zhao2025taste} and RT-1~\cite{brohan2022rt} datasets.  ``w/o ROI images'' denotes using only the environment image $\envimage$ as input. ``w/o local attention'' denotes removing the local-attention module.}
\label{tab:quant-last-frame}
\end{table}

\subsection{Experimental Results}
\paragraph{Goal image generation results.}
We evaluate goal-image generation on two datasets, RT1~\cite{brohan2022rt} and Taste-Rob~\cite{zhao2025taste}. For a fair comparison, all baseline predictions are resized to the ground-truth resolution. As shown in Table~\ref{tab:quant-last-frame}, our method consistently outperforms Nano Banana~\cite{google2025nanobanana} and Flux Kontext~\cite{batifol2025flux} across all metrics on both datasets, demonstrating clear improvements in goal-image quality.

\begin{figure}
  \centering
  \includegraphics[width=1.0\linewidth]{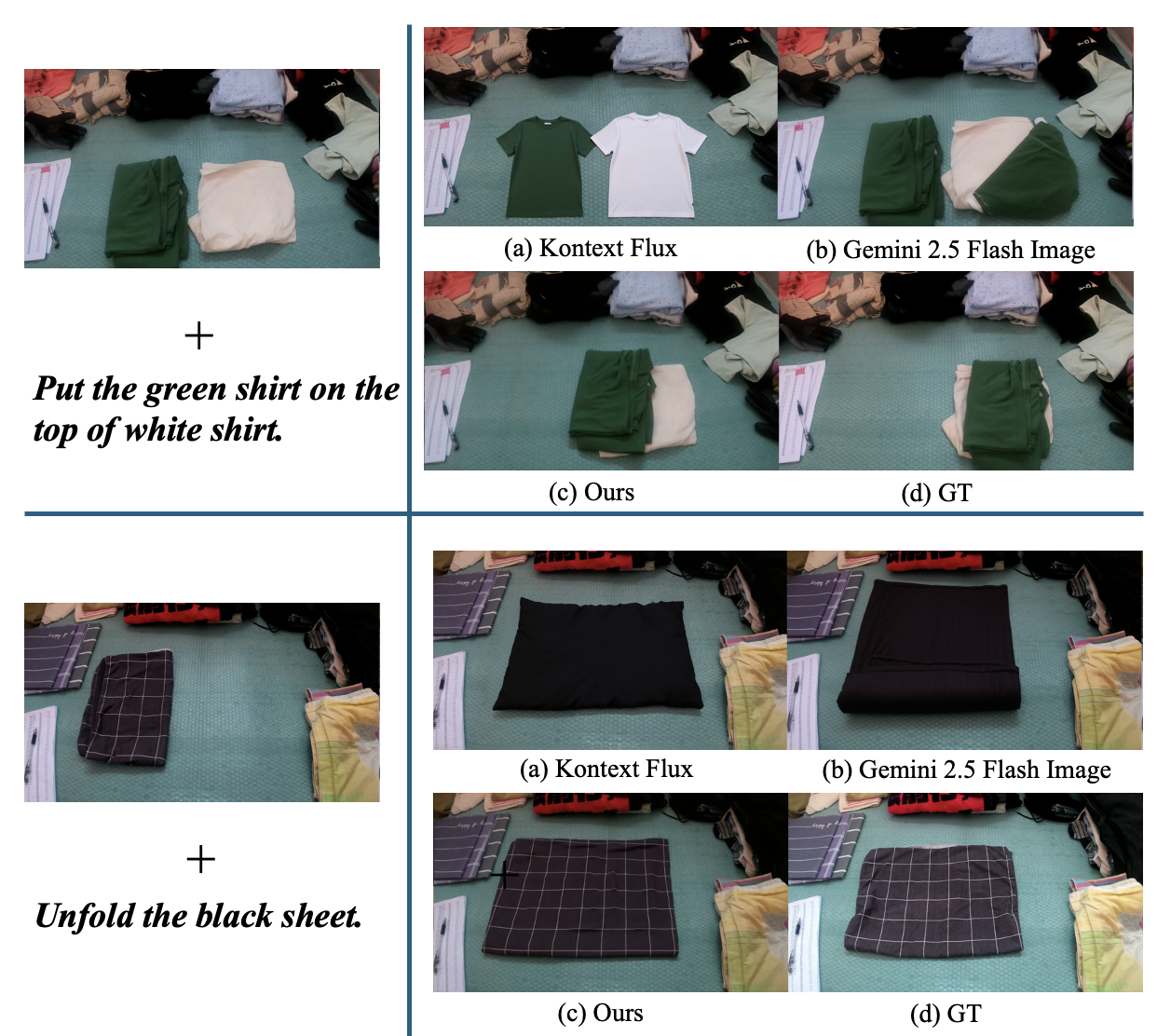}
\caption{Qualitative comparison of goal-image generation on Taste-Rob~\cite{zhao2025taste}. Our method follows instructions effectively and produces goal images with the highest perceptual quality while accurately preserving object shapes. Even for highly deformable objects with significant articulation and complex geometry (such as black sheets and green shirts), our approach maintains coherent geometry and preserves structural integrity.}
  \label{fig:goal_image_comparison}
\end{figure}

To assess the contribution of each proposed component, we perform ablation studies. As shown in Figure~\ref{fig:goal_image_ablation}, removing either the enhanced local image attention or the local text attention leads to unintended edits and spurious objects, as the model misinterprets the correspondence between prompts and image regions. Table~\ref{tab:quant-last-frame} and Figure~\ref{fig:goal_image_comparison} consistently show that incorporating both ROI images and local attention yields significant performance gains. These results demonstrate that our designed components enable the backbone to more precisely localize task-relevant regions and reliably follow the given instructions.

\paragraph{Video planning results.}
As shown in Figures~\ref{fig:nanobanana_video_compare}, \ref{fig:cross_embodiement}, and~\ref{fig:comparison_with_sota_video_mdoels}, our LoRA-finetuned Env-Goal Video Model produces planning videos with strong object-shape consistency, high perceptual quality, physical plausibility, and accurate instruction following. We attribute this to the FL2V architecture (Section~\ref{sec:fl2v_video_model}), which explicitly preserves the goal state. In particular, Figure~\ref{fig:nanobanana_video_compare} shows that Nanobanana~\cite{google2025nanobanana} fails to generate a reasonable final frame for downstream video generation, while our goal imagery model yields physically plausible and semantically consistent goal frames that effectively guide the video synthesis process.

We compare against state-of-the-art video generation/planning methods (Tesseract~\cite{zhen2025tesseract}, Taste-Rob-I2V~\cite{zhao2025taste}, and Wan-2.1-I2V~\cite{wan2025wan}) on Taste-Rob~\cite{zhao2025taste} and RT1~\cite{brohan2022rt} datasets. 
Taste-Rob-I2V and Tesseract are originally pre-trained on Taste-Rob and RT1, respectively.
For a fair comparison, we fine-tune Wan-2.1-I2V on each dataset using LoRA~\cite{hu2021lora} and denote the adapted version as Wan-2.1-I2V*.
As shown in Figure~\ref{fig:comparison_with_sota_video_mdoels}, Wan-2.1-I2V* frequently fails to follow the textual instructions, while Tesseract and Taste-Rob-I2V struggle to preserve the shapes of both manipulated and static objects. In contrast, our method produces physically plausible object motions and faithfully follows the prompt.
We further compare all approaches using FVD, LPIPS, SSIM (video quality), PA (physical alignment), and IF (instruction following). As reported in Table~\ref{tab:quant-video-table}, our model achieves significant gains across all metrics.
Finally, Figure~\ref{fig:cross_embodiement} demonstrates that training on heterogeneous datasets enables strong cross-domain generalization: our method handles both robot and human hands and generalizes to unseen embodiments, supporting applications such as large-scale data augmentation for robot planning.

\paragraph{Robotic planning results.}
We evaluate on a mixed dataset comprising six self-collected challenging tasks in IsaacGym~\cite{makoviychuk2021isaac} (Table~\ref{tab:quant-robot_rollout}) and five tasks from Robotmimic~\cite{mandlekar2021matters}. For the IsaacGym experiments, we use an Allegro hand~\cite{allegro} paired with an XArm~\cite{githubGitHubXArmDeveloperxarm_ros2} system.
Additional setup details are provided in the supplementary material.

We compare our method with state-of-the-art video planning baselines, \ie, Tesseract~\cite{zhen2025tesseract} and Wan-2.1-I2V~\cite{wan2025wan}, which we fine-tune on the same data for fair comparison. We first generate planning videos and then extract actions from RGB frames using Diffusion Policy~\cite{chi2025diffusion} for execution.
As shown in Figure~\ref{fig: rollbot simulation} and Table~\ref{tab:quant-robot_rollout}, methods that rely purely on forward image-to-video prediction, such as Tesseract, often introduce object distortions over time. These deformations corrupt the visual frames and lead Diffusion Policy to infer incorrect control signals, resulting in lower success rates. In contrast, \sysname generates more stable and physically consistent videos: by conditioning on both the initial and goal observations, it constrains the temporal evolution and preserves object shape throughout the sequence.

\section{Conclusion}
In this work, we present \sysname, a goal-conditioned visual planning framework that predicts a coherent goal image and then generates a physically consistent trajectory connecting the start and end states. By explicitly predicting and conditioning on the desired end state, and by using local attention to precisely ground text instructions to task-relevant regions, \sysname mitigates the spatial drift, object deformation, and ambiguity common in forward-only baselines. This design leads to substantial improvements in goal image fidelity, temporal coherence, and downstream robot execution across both human-hand and robot-arm domains. While conditioning on a goal frame substantially enhances physical reliability, future work can further advance goal image reasoning, integrate richer 3D cues, and extend the framework to longer-horizon tasks.




{
    \small
    \bibliographystyle{ieeenat_fullname}
    \bibliography{main}
}


\end{document}